\documentclass[10pt,twocolumn,letterpaper]{article}

\usepackage[pagenumbers]{cvpr}      

\usepackage{graphicx}
\usepackage{amsmath}
\usepackage{amssymb}
\usepackage{booktabs}
\usepackage{algorithm}
\usepackage{algorithmic}

\usepackage{graphicx, amsmath, amssymb, caption, subcaption, multirow, overpic, textpos}
\usepackage[table]{xcolor}
\usepackage[british, english, american]{babel}
\usepackage[pagebackref=false, breaklinks=true, letterpaper=true, colorlinks,citecolor=citecolor, linkcolor=linkcolor, bookmarks=false]{hyperref}
\usepackage{paralist}
\definecolor{citecolor}{HTML}{0000FF}
\definecolor{linkcolor}{HTML}{FF0000}
\usepackage{xcolor}
\definecolor{ColorName}{RGB}{230,30,70}
\newlength\savewidth\newcommand\shline{\noalign{\global\savewidth\arrayrulewidth
\global\arrayrulewidth 1pt}\hline\noalign{\global\arrayrulewidth\savewidth}}
\newcommand{\tablestyle}[2]{\setlength{\tabcolsep}{#1}\renewcommand{\arraystretch}{#2}\centering\footnotesize}

\newcommand{\app}{\raise.17ex\hbox{$\scriptstyle\sim$}}

\definecolor{deemph}{gray}{0.6}
\newcommand{\gc}[1]{\textcolor{deemph}{#1}}
\definecolor{baselinecolor}{gray}{.9}

\usepackage{amsmath}
\usepackage{amssymb}
\usepackage{epsfig}
\usepackage{graphicx}
\usepackage{booktabs}
\usepackage{bbding}
\usepackage{pifont}
\usepackage[capitalize]{cleveref}
\crefname{section}{Sec.}{Secs.}
\Crefname{section}{Section}{Sections}
\Crefname{table}{Table}{Tables}
\crefname{table}{Tab.}{Tabs.}

\def\cvprPaperID{1182} 
\def\confName{ICCV}
\def\confYear{2023}

\begin{document}

\title{SelfPromer: Self-Prompt Dehazing Transformers with Depth-Consistency}

\author{Cong Wang$^{1}$, Jinshan Pan$^{2}$, Wanyu Lin$^{1}$, Jiangxin Dong$^{2}$, Xiao-Ming Wu$^{1}$ \\
$^{1}$The Hong Kong Polytechnic University, $^{2}$Nanjing University of Science and Technology\\
{\tt\small \{supercong94;sdluran;dongjxjx\}@gmail.com, \{wan-yu.lin;xiao-ming.wu\}@polyu.edu.hk}
}
\maketitle

\begin{abstract}

This work presents an effective depth-consistency self-prompt Transformer for image dehazing. It is motivated by an observation that the estimated depths of an image with haze residuals and its clear counterpart vary. Enforcing the depth consistency of dehazed images with clear ones, therefore, is essential for dehazing. For this purpose, we develop a prompt based on the features of depth differences between the hazy input images and corresponding clear counterparts that can guide dehazing models for better restoration. Specifically, we first apply deep features extracted from the input images to the depth difference features for generating the prompt that contains the haze residual information in the input. Then we propose a prompt embedding module that is designed to perceive the haze residuals, by linearly adding the prompt to the deep features. Further, we develop an effective prompt attention module to pay more attention to haze residuals for better removal. By incorporating the prompt, prompt embedding, and prompt attention into an encoder-decoder network based on VQGAN, we can achieve better perception quality. As the depths of clear images are not available at inference, and the dehazed images with one-time feed-forward execution may still contain a portion of haze residuals, we propose a new continuous self-prompt inference that can iteratively correct the dehazing model towards better haze-free image generation. Extensive experiments show that our method performs favorably against the state-of-the-art approaches on both synthetic and real-world datasets in terms of perception metrics including NIQE, PI, and PIQE.


\end{abstract}

\vspace{-3mm}
\section{Introduction}\label{sec:intro}
%
Recent years have witnessed advanced progress in image dehazing due to the development of deep dehazing models.
Mathematically, the haze process is usually modeled by an atmospheric light scattering model~\cite{dehazing_tpami11_darkchannel} formulated as:
\begin{small}
\begin{equation}
\text{I}(x) = \text{J}(x)\text{T}(x) + (1-\text{T}(x))\text{A},
\label{eq:hazy formation}
\end{equation}
\end{small}
where $\text{I}$ and $\text{J}$ denote a hazy and haze-free image, respectively, and $\text{A}$ denotes the global atmospheric light, $x$ denotes the pixel index, and the transmission map $\text{T}$ is usually modeled as $\text{T}(x) = e^{-\beta \text{d}(x)}$ with the scene depth $\text{d}(x)$, and the scattering coefficient $\beta$ reflects the haze density.
Most existing works develop various variations of deep Convolutional Neural Networks (CNNs) for image dehazing~\cite{Dehazing-semi-li-tip20,msbdn_cvpr20_dong,kddn_dehaze_cvpr,pfdn_eccv20_dong,aecr-cvpr21-wu,guo2022dehamer}.
They typically compute a sequence of features from the hazy input images and directly reconstruct the clear ones based on the features, which have achieved state-of-the-art results on benchmarks~\cite{RESIDE_dehazingbenchmarking_tip2019} in terms of PSNRs and SSIMs.
However, as dehazing is ill-posed, very small errors in the estimated features may degrade the performance.
Existing works propose to use deep CNNs as image priors and then restore the clear images iteratively. However, they cannot effectively correct the errors or remove the haze residuals in the dehazed images as these models are fixed in the iterative process~\cite{Liu_2019_ICCV}.
It is noteworthy that the human visual system generally possesses an intrinsic correction mechanism that aids in ensuring optimal results for a task. This phenomenon has been a key inspiration behind the development of a novel dehazing approach incorporating a correction mechanism that guides deep models toward better haze-free results generation.

Specifically, if a dehazed result exists haze residuals, a correction mechanism can localize these regions and guide the relevant task toward removing them.
Notably, NLP-based text prompt learning has shown promise in guiding the models by correcting the predictions~\cite{liu2023pre}.
However, text-based prompts may not be appropriate for tasks that require solely visual inputs without accompanying text.
Recent works~\cite{PromptonomyViT,Decorate} attempted to address this issue by introducing text-free prompts into vision tasks.
For instance, PromptonomyViT~\cite{PromptonomyViT} evaluates the adaptation of multi-task prompts such as depth, normal, and segmentation to improve the performance of the video Transformers.
Nevertheless, these prompts may not be suitable for image dehazing tasks, as they could not capture the haze-related content.

\begin{figure*}[t]
\vspace{-2mm}
\footnotesize
\centering
\begin{center}
\begin{tabular}{ccccccccc}
\includegraphics[width=0.99999999\linewidth]{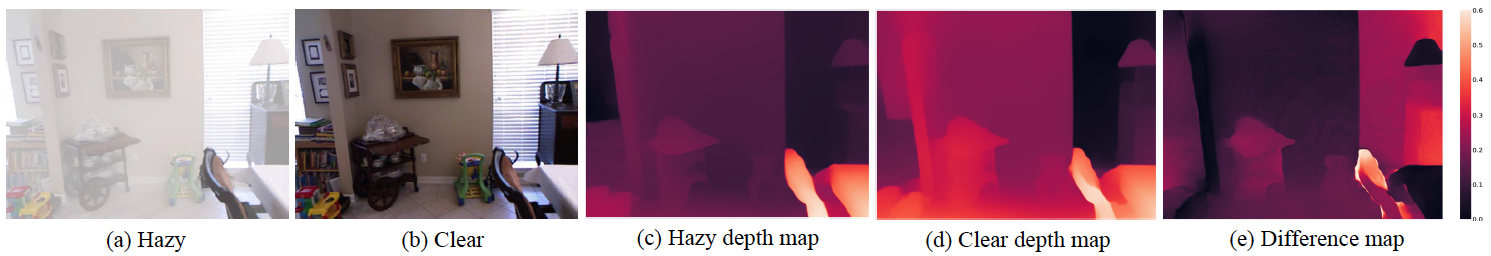}
\\
\end{tabular}
\vspace{-5mm}
\caption{
Haze residuals pose a significant challenge to accurately estimating the depth of clear images, creating inconsistencies compared to hazy images.
A difference map (e) is utilized to locate haze residuals on the estimated depth, while minimal haze residuals will result in consistent estimates.
By analyzing the difference map, we can identify the impact of haze residuals, leading to the development of improved dehazing models to mitigate this effect and enhance the quality of dehazed images.
The difference map (e) is derived by $|\text{hazy~depth}-\text{clear~depth}|$ with equalization for better visualization.
}
\label{fig:Visualization on feature-level depth difference}
\end{center}
\vspace{-8mm}
\end{figure*}

To better guide the deep model for better image dehazing, this work develops an effective self-prompt dehazing Transformer. Specifically, it explores with the depth consistency of hazy images and their corresponding clear ones as a prompt.
In particular, our study is motivated by the substantial difference between the estimated depths of hazy images and their clear counterparts, \ie, the same scene captured in the same location should be consistent regarding depth.
Depth is typically related to the transmission map in the atmospheric light scattering model as shown in~\eqref{eq:hazy formation}.
Thus, if the dehazed images can be reconstructed accurately, their estimated depths should be close to those of their clear counterparts at large.
However, haze residuals often degrade the accuracy of depth estimation, resulting in significant differences between hazy and clear images, as illustrated in Fig.~\ref{fig:Visualization on feature-level depth difference}(e).
Yet, the difference map of estimated depths from images with haze residuals and clear images often points to the regions affected by haze residuals.

Based on the above observation, we design a prompt to guide the deep models for perceiving and paying more attention to haze residuals.
Our prompt is built upon the estimated feature-level depth differences, of which the inconsistent regions can reveal haze residual locations for deep models correction.
On top of the prompt, we introduce a prompt embedding module that linearly combines input features with the prompt to better perceive haze residuals.
Further, we propose a prompt attention module that employs self-attention guided by the prompt to pay more attention to haze residuals for better haze removal.
Our encoder-decoder architecture combines these modules using VQGAN~\cite{vqgan} to enhance the perception quality of the results, as opposed to relying solely on PSNRs and SSIMs metrics for evaluation.
As the depths of clear images suffer from unavailability at inference and dehazed images obtained via one-time feed-forward execution may have haze residuals, we introduce a continuous self-prompt inference to address these challenges.
Specifically, our proposed approach feeds the hazy input image to the model and sets the depth difference as zero to generate clearer images that serve as the clear counterpart.
The clear image participates in constructing the prompt to conduct prompt dehazing.
The inference operation is continuously conducted as the depth differences can keep correcting the deep dehazing models toward better haze-free image generation.

This paper makes the following contributions:
\begin{compactitem}
\item We make the first attempt to formulate the prompt by considering the cues of the estimated depth differences between the image with haze residuals and its clear counterpart in the image dehazing task.


\item We propose a prompt embedding module and a prompt attention module to respectively perceive and pay more attention to haze residuals for better removal.
\item We propose a new continuous self-prompt inference approach to iteratively correct the deep models toward better haze-free image generation.

\item
Experiments demonstrate that our method performs favorably against state-of-the-art approaches on both synthetic and real-world datasets in terms of perception metrics including NIQE, PI, and PIQE.
\end{compactitem}

%
\begin{figure*}[!t]
\vspace{-2mm}
\footnotesize
\centering
\begin{center}
\begin{tabular}{c}
\includegraphics[width=0.99\linewidth]{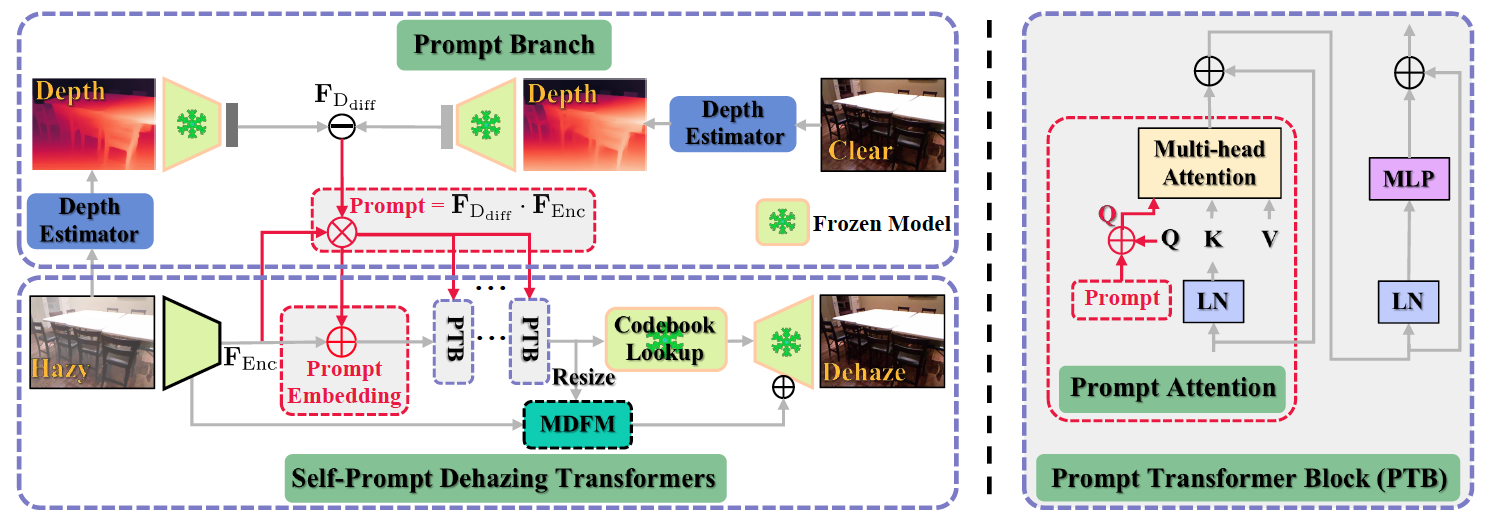}
\end{tabular}
\vspace{-5mm}
\caption{\textbf{The proposed framework at training stage}.
MDFM is detailed in Sec.~\ref{sec:Multual deformable fusion module}.
The inference process is illustrated in Fig.~\ref{fig: Continuous Depth-Consistency Self-Prompt Transformers at testing stage}.
}
\label{fig:Continuous Depth-Consistency Self-Prompt Transformers at training stage}
\end{center}
\vspace{-8mm}
\end{figure*}
\section{Related Work}
In this section, we overview image dehazing, VQGAN image restoration, and prompt vision applications.
%
%
%
\\
{\bf Image Dehazing.}
Traditional solutions usually design various hand-crafted priors captured deterministic and statistical properties of hazy and haze-free images to remove haze, such as dark channel~\cite{dehazing_tpami11_darkchannel}, color-line~\cite{dehazing_TOG14_colorline}, haze-line~\cite{dehazing_iccp17_hazelines}, \etc.
Recently, CNN-based dehazing approaches are gradually developed~\cite{dehazing_tip16_dehazenet,dehazing_eccv16_mscnn}, \eg, MSCNN~\cite{dehazing_eccv16_mscnn} use CNN to estimate the transmission map.
One limitation of these algorithms is not flexible as they are not end-to-end.
To address this issue, end-to-end dehazing networks~\cite{dehazing_AOD,pfdn_eccv20_dong,msbdn_cvpr20_dong,dehazing_cvpr19,grid_dehaze_liu} are proposed.
Considering the haze physics model~\eqref{eq:hazy formation}, physics-based CNNs~\cite{dehazing_tcsvt19_zhang,dcpdn,dualcnn_cvpr18,pan2022dual_ijcv} are suggested.
Motivated the powerful generation ability of CycleGAN~\cite{cyclegan}, cycle-based methods~\cite{physicsgan_pan,d4_dehze} are adapted.
Although these efforts, these methods usually tend to produce unsatisfactory results as they cannot effectively perceive haze residuals.
\\
{\bf VQGAN for Image Restoration.}
Recent research~\cite{vqfr, RestoreFormer,codeformer,mm22vqsr} has shown that VQGAN~\cite{vqgan} is an effective tool to generate more realistic results.
VQGAN-based restoration methods estimate latent clear images but often neglect deep model prior cues, which can limit their performance.
Zhou~\etal. propose CodeFormer~\cite{codeformer}, which inserts regular Transformers into VQGAN for face restoration.
Different from this work, our approach incorporates the estimated depth inconsistency between the image with haze residuals and its clear version by using prompt embedding and prompt attention to iteratively correct deep models with a self-prompt inference scheme for image dehazing.
%
\\
{\bf Prompt Learning for Vision.}
Prompt learning is first studied in natural language processing~\cite{prompt_Exploiting,AutoPrompt}.
Due to its high effectiveness, prompt learning is recently used in vision-related tasks~\cite{Conditional_Prompt_Learning,CPT,Unified,Open_Vocabulary}, \eg, domain generalization~\cite{Prompt_Vision_Transformer}, multi-modal learning~\cite{MaPLe}, action understanding~\cite{Bridge_Prompt}, and visual prompt tuning~\cite{vpt}.
To our knowledge, there is no effort to exploit prompts for dehazing. This paper aims to investigate this new path.
\section{Proposed Method}
Our method comprises two branches: the prompt branch and the self-prompt dehazing Transformer branch. The prompt branch generates a prompt by using the deep depth difference and deep feature extracted from the hazy input.
The other branch exploits the generated prompt to guide the deep model for image dehazing.
We incorporate a prompt embedding module and prompt attention module to respectively perceive and pay more attention to the haze residuals for better removal. The proposed modules are formulated into an encoder-decoder architecture based on VQGAN for better perception quality~\cite{codeformer,mm22vqsr,vqfr}.
%

%
\subsection{Overall Framework}\label{sec:Overall Framework}
Fig.~\ref{fig:Continuous Depth-Consistency Self-Prompt Transformers at training stage} illustrates our method at the training stage.
Given a hazy images $\text{I}$, we first utilize trainable encoder $\textit{Enc}(\cdot)$ to extract features:
\begin{small}
\vspace{-1mm}
\begin{equation}\label{eq:spatially-reduced features}
\mathbf{F}_{\text{Enc}} = \textit{Enc}(\text{I}).
\end{equation}
\end{small}

Then, we compute the depth difference of the hazy image $\text{I}$ and its corresponding clear image $\text{J}$ in feature space: 
\begin{small}
\begin{subequations}\label{eq:feature-level depth difference}
\vspace{-2mm}
\begin{align}
&\text{D}_{1} = \textit{DE}(\text{I});~\text{D}_{2} = \textit{DE}(\text{J}),\\
&\mathbf{F}_{\text{D}_{1}} = \textit{Enc}_{\text{pre}}^{\text{frozen}}(\text{D}_{1});~\mathbf{F}_{\text{D}_{2}} = \textit{Enc}_{\text{pre}}^{\text{frozen}}(\text{D}_{2}),\\
&\mathbf{F}_{\text{D}_{\text{diff}}} = |\mathbf{F}_{\text{D}_{1}} -\mathbf{F}_{\text{D}_{2}}|,
\end{align}
\end{subequations}
\end{small}
\\
\vspace{-10mm}
\\
where $\textit{DE}(\cdot)$ denotes the depth estimator\footnote{We chose DPT\_Next\_ViT\_L\_384 to balance accuracy, speed, and model size: https://github.com/isl-org/MiDaS.}~\cite{Ranftl2020}.
$\textit{Enc}_{\text{pre}}^{\text{frozen}}(\cdot)$ denotes the pre-trained VQGAN encoder which is frozen when training our dehazing models.

Next, we exploit $\mathbf{F}_{\text{D}_{\text{diff}}}$ to build the $\mathrm{Prompt}$, and develop a prompt embedding module and a prompt attention module in Transformers, \ie, $\textit{PTB}$ (see details in Sec.~\ref{sec:Depth-consistency self-prompt Transformers}) to better generate haze-aware features:
\begin{small}
\begin{subequations}\label{eq:feature-level depth difference}
\vspace{-1mm}
\begin{align}
&\mathrm{Prompt} = \mathbf{F}_{\text{D}_{\text{diff}}} \cdot \mathbf{F}_{\text{Enc}},~~~~~~~~~~~~~~~~~\textcolor{blue}{\text{\#~Prompt}}~\label{eq:Prompt}
\\
&\mathbf{F}_{\text{ProEmbed}} = \mathrm{Prompt} + \mathbf{F}_{\text{Enc}},~~~~~~~~~\textcolor{blue}{\text{\#~Prompt~Embedding}}\label{eq:Prompt Embedding}
\\
&\mathbf{F}_{\text{PTB}} = \textit{PTB}(\mathrm{Prompt}, \mathbf{F}_{\text{ProEmbed}}),~\textcolor{blue}{\text{\#~Prompt~Transformer}}
\end{align}
\end{subequations}
\end{small}
\\
\vspace{-10mm}
\\
where $\mathbf{F}_{\text{ProEmbed}}$ means the features of prompt embedding.

\begin{figure}[!t]
\vspace{-2mm}
\footnotesize
\centering
\begin{center}
\begin{tabular}{c}
 \hspace{-4mm}\includegraphics[width=1.04\linewidth]{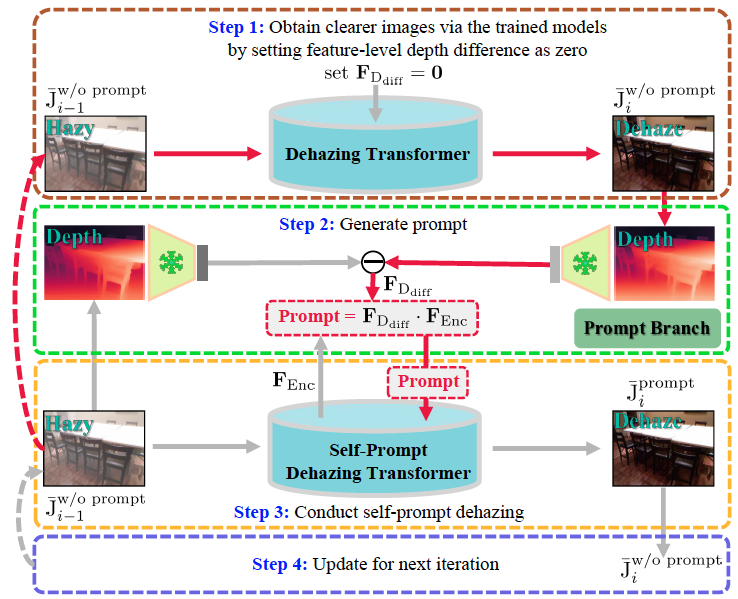}
\end{tabular}
\vspace{-4mm}
\caption{\textbf{Continuous Self-Prompt Inference}.
$i^{\text{th}}$ prompt inference contains \textcolor{blue}{\text{four steps}}: Sequential execution from top to bottom.
Step~1 obtains clearer images to participate in forming the prompt by feeding the hazy image itself to our network without prompt by setting $\mathbf{F}_{\text{D}_{\text{diff}}}$ as zero.
Step~2 generates the prompt to guide the dehazing model.
Step~3 conducts the self-prompt dehazing to produce the results.
Step~4 updates for the next iterative dehazing.
The {\color{ColorName}{\textbf{magenta line}}} describes the 'self' process that builds the prompt from the hazy image itself.
Here, Dehazing Transformer means our Self-Prompt Dehazing Transformer with $\mathbf{F}_{\text{D}_{\text{diff}}}=0$.
}
\label{fig: Continuous Depth-Consistency Self-Prompt Transformers at testing stage}
\end{center}
\vspace{-8mm}
\end{figure}

The generated feature $\mathbf{F}_{\text{PTB}}$ is further matched with the learned haze-free $\textbf{\texttt{Codebook}}$ at the pre-trained VQGAN stage by the $\textit{Lookup}$ method~\cite{vqgan,codeformer}:
\begin{small}
\vspace{-1mm}
\begin{equation}\label{eq:Lookup}
\mathbf{F}_{\text{mat}} = \textit{Lookup}(\mathbf{F}_{\text{PTB}}, \textbf{\texttt{Codebook}}).
\end{equation}
\end{small}
\\
\vspace{-14mm}
\\

Finally, we reconstruct the dehazing images $\Bar{\text{J}}$ from the matched features $\mathbf{F}_{\text{mat}}$ by decoder of pre-trained VQGAN $\textit{Dec}_{\text{pre}}^{\text{frozen}}(\cdot)$ with residual learning~\cite{mm22vqsr} by mutual deformable fusion module $\textit{MDFM}$ (see details in Sec.~\ref{sec:Multual deformable fusion module}):
\begin{small}
\vspace{-1mm}
\begin{equation}
\label{eq:finaloutput}
\Bar{\text{J}} = \textit{Dec}_{\text{pre}}^{\text{frozen}}(\mathbf{F}_{\text{mat}}) + \textit{MDFM}\Big(\textbf{F}_{\text{Enc}}^{\text{s}}, \textbf{F}_{\text{PTB}}^{\text{s,u}}\Big),
\end{equation}
\end{small}
\\
\vspace{-10mm}
\\
where $\textbf{F}_{\text{Enc}}^{\text{s}}$ means the encoder features at $\text{s}$ scale, while $\textbf{F}_{\text{PTB}}^{\text{s,u}}$ denotes the $\text{s}\times$ upsampling features of $\textit{PTB}$.
We conduct the residual learning with MDFM in $\{1, 1/2, 1/4, 1/8\}$ scales between the encoder and decoder like FeMaSR~\cite{mm22vqsr}.
Here, $\textbf{F}_{\text{Enc}}^{\text{1/8}}$ denotes the $\textbf{F}_{\text{Enc}}$ in \eqref{eq:spatially-reduced features}.
\\
\textbf{Loss Functions.}
We use pixel reconstruction loss $\mathcal{L}_{\text{rec}}$, codebook loss $\mathcal{L}_{\text{code}}$, perception loss $\mathcal{L}_{\text{per}}$, and adversarial loss $\mathcal{L}_{\text{adv}}$ to measure the error between the dehazed images $\Bar{\text{J}}$ and the corresponding ground truth $\text{J}$:
\begin{small}
\vspace{-2mm}
\begin{equation}
\begin{array}{ll}
\mathcal{L} =\mathcal{L}_{\text{rec}}+\lambda_{\text{code}}\mathcal{L}_{\text{code}} + \lambda_{\text{per}}\mathcal{L}_{\text{per}} + \lambda_{\text{adv}}\mathcal{L}_{\text{adv}},
\end{array}
\label{eq:pw-stb}
\end{equation}
\end{small}
\\
\vspace{-10mm}
\\
where
\begin{small}
\vspace{-2mm}
\begin{subequations}
\label{eq:pw-stb}
\begin{align}
&\mathcal{L}_{\text{rec}} =||\Bar{\text{J}}-\text{J}||_{1} + \lambda_{\text{ssim}}\big(1-\textit{SSIM}(\Bar{\text{J}},\text{J})\big),\\
& \mathcal{L}_{\text{code}} = ||\bar{z}_{\mathbf{q}}-z_{\mathbf{q}}||_{2}^{2},\\
&\mathcal{L}_{\text{per}} = ||\Phi(\Bar{\text{J}})- \Phi(\text{J})||_{2}^{2},\\
&\mathcal{L}_{\text{adv}}= \mathbb{E}_{\text{J}}[\mathrm{log}~\mathcal{D}(\text{J})] + \mathbb{E}_{\Bar{\text{J}}}[1-\mathrm{log}~\mathcal{D}(\Bar{\text{J}})],
\end{align}
\end{subequations}
\end{small}
\\
\vspace{-10mm}
\\
where $\textit{SSIM}(\cdot)$ denotes the structural similarity~\cite{SSIM_wang} for better structure generation.
${z}_{\mathbf{q}}$ is the haze-free codebook features by feeding haze-free images $\text{J}$ to pre-trained VQGAN while $\bar{z}_{\mathbf{q}}$ is the reconstructed codebook features.
$\Phi(\cdot)$ denotes the feature extractor of VGG19~\cite{vgg19}.
$\mathcal{D}$ is the discriminator~\cite{cyclegan}.
$\lambda_{\text{code}}, \lambda_{\text{per}}, \lambda_{\text{adv}}$, and $\lambda_{\text{ssim}}$ are weights.

For {\bf inference}, we propose a new self-prompt inference approach (see details in Sec.~\ref{sec:Inference with depth-consistency self-prompt}) as our training stage involves the depth of clear images to participate in forming the prompt while clear images are not available at testing.
\subsection{Self-Prompt Transformers}\label{sec:Depth-consistency self-prompt Transformers}
%
%

The proposed self-prompt Transformer contains the prompt generated by the prompt branch, a prompt embedding module, and a prompt attention module which is contained in the prompt Transformer block.
In the following, we introduce the definition of the prompt, prompt embedding module and prompt attention module, and prompt Transformer block in detail.
\\
\textbf{Prompt} (Definition).
The prompt is based on the estimated depth difference between the input image and its clear counterpart.
It is defined in \eqref{eq:Prompt} which can better contain haze residual features as $\mathbf{F}_{\text{D}_{\text{diff}}}$ with higher response value reveals inconsistent parts which potentially correspond to the haze residuals in the input hazy image.
\begin{figure}[!t]
\vspace{-2mm}
\footnotesize
\centering
\begin{center}
\begin{tabular}{c}
\includegraphics[width=0.95\linewidth]{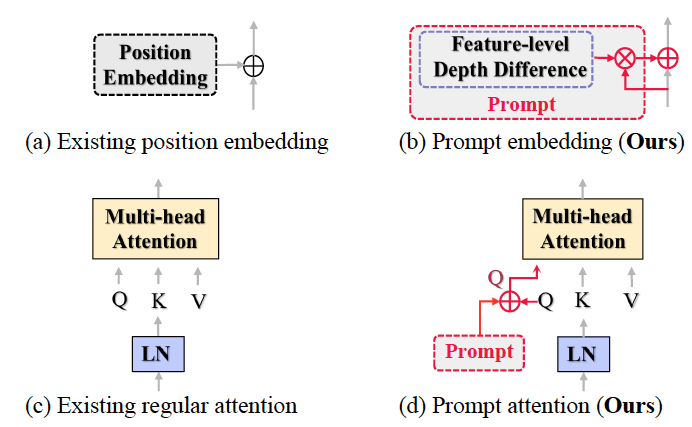}
\end{tabular}
\vspace{-5mm}
\caption{\textbf{(a)-(b) Existing position embedding \vs Prompt embedding (Ours)}.
Our prompt embedding can better perceive the haze information and is friendly for different input sizes.
\textbf{(c)-(d) Existing regular attention \vs Prompt attention (Ours)}.
Our prompt attention can pay more attention to the haze residuals.}
\label{fig:Existing attention Prompt-based attention}
\end{center}
\vspace{-8mm}
\end{figure}
\\\textbf{Prompt Embedding.}
Existing Transformers~\cite{Prompt_Vision_Transformer} usually use the position embedding method (Fig.~\ref{fig:Existing attention Prompt-based attention}(a)) to represent the positional correlation, which does not contain haze-related information so that it may not effectively perceive the haze residual information well.
Moreover, image restoration requires processing different input sizes at inference while the position embedding is defined with fixed parameters at training~\cite{Prompt_Vision_Transformer}.
Hence, position embedding may be not a good choice for image dehazing.
To overcome these problems, we propose prompt embedding which is defined in \eqref{eq:Prompt Embedding}.
By linearly adding the extracted features $\mathbf{F}_{\text{Enc}}$ with $\mathrm{Prompt}$, the embedded feature $\mathbf{F}_{\text{ProEmbed}}$ perceives the haze residual features as $\mathrm{Prompt}$ extracts the haze residual features.
Note that as $\mathbf{F}_{\text{ProEmbed}}$ has the same size as $\mathbf{F}_{\text{Enc}}$, it does not require fixed sizes like position embedding.
\\
\textbf{Prompt Attention.}
Existing Transformers usually extract Query $\mathbf{Q}$, Key $\mathbf{K}$, and Value $\mathbf{V}$ from input features to estimate scaled-dot-product attention shown in Fig.~\ref{fig:Existing attention Prompt-based attention}(c).
Although Transformers are effective for feature representation, the standard operation may be not suitable for image dehazing.
To ensure the Transformers pay more attention to haze residuals for better removal, we propose prompt attention $\textit{ProAtt}(\cdot)$ by linearly adding the query with $\mathrm{Prompt}$:
%
%
\begin{small}
\vspace{-5mm}
\begin{subequations}\label{eq:ProAtt}
\begin{align}
&\textcolor{red}{\textbf{Q}} =  \textbf{Q} + \mathrm{Prompt},
\\
&\textit{ProAtt}(\textbf{Q}, \textbf{K}, \textbf{V}) = \textit{Softmax}\Big(\frac{\textcolor{red}{\textbf{Q}}\textbf{K}^{\text{T}}}{\sqrt{d_{\text{head}}}}\Big)\textbf{V},
\end{align}
\end{subequations}
\end{small}
\\
\vspace{-8mm}
\\
where $d_{\text{head}}$ denotes the dimension of head.
Fig.~\ref{fig:Existing attention Prompt-based attention}(d) illustrates the proposed prompt attention.
Note that as $\textbf{Q}$ in attention is to achieve the similarity relation for expected inputs~\cite{query}, our prompt attention by linearly adding the prompt $\mathrm{Prompt}$ with the Query $\textbf{Q}$ can pay more attention to haze residuals for better removal.
%
\\
\textbf{Prompt Transformer Block.}
According to the above attention design, our prompt Transformer block (PTB) can be sequentially computed as:
\begin{small}
\vspace{-2mm}
\begin{subequations}\label{eq:Prompt Transformer Block}
\begin{align}
& \textbf{Q}, \textbf{K}, \textbf{V} = \textit{LN}(\mathbf{X}^{l-1}), \\
&\hat{\mathbf{X}}^{l} =\textit{ProAtt}(\textbf{Q}, \textbf{K}, \textbf{V}) + \mathbf{X}^{l-1},
\\
&\mathbf{X}^{l} =  \textit{MLP}\Big(\textit{LN}(\hat{\mathbf{X}}^{l}\big)\Big) + \hat{\mathbf{X}}^{l},
\end{align}
\end{subequations}
\end{small}
\\
\vspace{-10mm}
\\
where $\mathbf{X}^{l-1}$ and $\mathbf{X}^{l}$ mean the input and output of the $l^{\text{th}}$ prompt Transformer block.
Specially, $\mathbf{X}^{0}$ is the $\mathbf{F}_{\text{ProEmbed}}$.
$\textit{LN}$ and $\textit{MLP}$ denote the layer normalization and multilayer perceptron.
The PTB is shown in the right part of Fig.~\ref{fig:Continuous Depth-Consistency Self-Prompt Transformers at training stage}.


It is worth noting that our prompt embedding and prompt attention are flexible as we can manually set $\mathbf{F}_{\text{D}_{\text{diff}}}=\mathbf{0}$, the network thus automatically degrade to the model without prompt, which will be exploited to form our continuous self-prompt inference (see Sec.~\ref{sec:Inference with depth-consistency self-prompt}).
\subsection{Mutual Deformable Fusion Module}\label{sec:Multual deformable fusion module}
As VQGAN is less effective for preserving details~\cite{vqfr,mm22vqsr}, motivated by the deformable models~\cite{Dai_2017_ICCV,Zhu_2019_CVPR_deformable} that can better fuse features, we propose a mutual deformable fusion module (MDFM) by fusing features mutually to adaptively learn more suitable offsets for better feature representation:
\begin{small}\label{eq:mutual deformable fusion module}
\vspace{-5mm}
\begin{subequations}
\begin{align}
&\hspace{-1.75mm}\text{off}_{1} = \textit{Conv}\Big(\mathcal{C}[\textbf{F}_{\text{Enc}}^{\text{s}}, \textbf{F}_{\text{PTB}}^{\text{s,u}}]\Big);\text{off}_{2} =  \textit{Conv}\Big(\mathcal{C}[\textbf{F}_{\text{PTB}}^{\text{s,u}}, \textbf{F}_{\text{Enc}}^{\text{s}}]\Big),\\
&\hspace{-1.75mm}\mathbf{Y}_{1} = \textit{DMC}(\textbf{F}_{\text{Enc}}^{\text{s}}, \text{off}_{1});\mathbf{Y}_{2} = \textit{DMC}(\textbf{F}_{\text{PTB}}^{\text{s,u}}, \text{off}_{2}),\\
&\hspace{-1.75mm}\mathbf{F}_{\text{MDFM}} = \textit{Conv}\Big(\mathcal{C}[\mathbf{Y}_{1}, \mathbf{Y}_{2}]\Big),
\end{align}
\end{subequations}
\end{small}
\\
\vspace{-10mm}
\\
where $\textit{Conv}(\cdot)$, $\mathcal{C}[\cdot]$, and $\textit{DMC}(\cdot)$ respectively denote the $1\times1$ convolution, concatenation, and deformable convolution.
$\text{off}_{k}~(k=1,2.)$ denotes the estimated offset.

\subsection{Continuous Self-Prompt Inference}\label{sec:Inference with depth-consistency self-prompt}
Our model requires the depth of clear images during training, but these images are unavailable at inference. Additionally, dehazed images generated by a one-time feed-forward execution may still contain some haze residuals. To address these issues, we propose a continuous self-prompt inference approach that leverages prompt embedding and prompt attention through linear addition, as discussed in Sec.~\ref{sec:Depth-consistency self-prompt Transformers}. By setting feature-level depth difference $\mathbf{F}_{\text{D}_{\text{diff}}}$ to zero, we can feed hazy images to our trained network and obtain clearer dehazed results which participate in building the prompt to conduct prompt dehazing.
The iterative inference is conducted to correct the deep models to ensure the deep models toward better haze-free image generation:
\begin{small}
\vspace{-2mm}
\begin{subequations}\label{eq: Self-prompt inference}
\begin{align}
&\hspace{-1.75mm}\Bar{\text{J}}_{i}^{\text{w/o~prompt}} = \mathcal{N}^{\text{w/o~prompt}}(\Bar{\text{J}}_{i-1}^{\text{w/o~prompt}}),~ \text{set}~\mathbf{F}_{\text{D}_{\text{diff}}}=\mathbf{0}, \hspace{-2.5mm}&\textcolor{blue}{\text{\#~Step 1}}\label{eq: Step 1}
\\
&\hspace{-1.75mm} \mathrm{Prompt} = \mathbf{F}_{\text{D}_{\text{diff}}} \cdot \mathbf{F}_{\text{Enc}};  \mathbf{F}_{\text{Enc}} = \textit{Enc}(\Bar{\text{J}}_{i-1}^{\text{w/o~prompt}}),\hspace{-2.5mm}&\textcolor{blue}{\text{\#~Step 2}}\label{eq: Step 2}
\\
&\hspace{-1.75mm}\Bar{\text{J}}_{i}^{\text{prompt}} = \mathcal{N}^{\text{prompt}}(\Bar{\text{J}}_{i-1}^{\text{w/o~prompt}},\mathrm{Prompt}),\hspace{-2.5mm}&\textcolor{blue}{\text{\#~Step 3}}\label{eq: Step 3}
\\
&\hspace{-1.75mm}\Bar{\text{J}}_{i}^{\text{w/o~prompt}} = \Bar{\text{J}}_{i}^{\text{prompt}},~~~(i=1,2, \cdots),\hspace{-2.5mm}&\textcolor{blue}{\text{\#~Step 4}}\label{eq: Step 4}
\end{align}
\end{subequations}
\end{small}
\\
\vspace{-10mm}
\\
where $\mathcal{N}^{\text{w/o~prompt}}$ denotes our trained network without prompt by setting $\mathbf{F}_{\text{D}_{\text{diff}}}$ as zero, while $\mathcal{N}^{\text{prompt}}$ means our trained network with prompt.
$\mathbf{F}_{\text{D}_{\text{diff}}} = |\textit{Enc}_{\text{pre}}^{\text{frozen}}(\textit{DE}(\Bar{\text{J}}_{i-1}^{\text{w/o~prompt}})) -\textit{Enc}_{\text{pre}}^{\text{frozen}}(\textit{DE}(\Bar{\text{J}}_{i}^{\text{w/o~prompt}}))|$.
$\Bar{\text{J}}_{0}^{\text{w/o~prompt}}$ denotes the original hazy images, while $\Bar{\text{J}}_{i-1}^{\text{w/o~prompt}}$ is regarded as the image with haze residuals and $\Bar{\text{J}}_{i}^{\text{w/o~prompt}}$ in \eqref{eq: Step 1} is regarded as the clear counterpart of $\Bar{\text{J}}_{i}^{\text{w/o~prompt}}$.
$\Bar{\text{J}}_{i}^{\text{prompt}}$ means the $i^{\text{th}}$ prompt dehazing results.

According to \eqref{eq: Self-prompt inference}, the inference is a \textbf{continuous self-prompt} scheme, \ie, we get the clear images from the hazy image itself by feeding it to $\mathcal{N}^{\text{w/o~prompt}}$ to participate in producing the prompt and the inference is continuously conducted.
Fig.~\ref{fig: Continuous Depth-Consistency Self-Prompt Transformers at testing stage} better illustrates the inference process.

\begin{figure}[!t]
\vspace{-1mm}
\footnotesize
\centering
\begin{center}
\begin{tabular}{c}
\hspace{-2mm}\includegraphics[width=0.999999\linewidth]{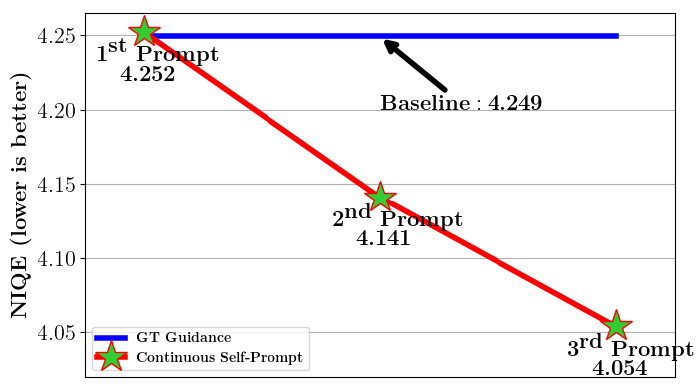}
\end{tabular}
\vspace{-4mm}
\caption{\textbf{Continuous self-prompt inference \vs. GT guidance (Baseline)} on the SOTS-indoor dataset.
GT guidance means we use the GT image to participate in forming the prompt at inference like the process of the training stage, which serves as the baseline.
Due to the one-time feed-forward execution that may still contain a portion of haze residuals, continuously conducting inference can ensure the results toward better haze-free image generation.
Note GT guidance only conducts one-time inference to generate a result.
What's more, GT is not available in the real world.
More detailed explanations are given in Sec.~\ref{sec:Analysis and Discussion}.
}
\label{fig: Continuous self-prompt vs GT guidance.}
\end{center}
\vspace{-8mm}
\end{figure}

Fig.~\ref{fig: Continuous self-prompt vs GT guidance.} shows our continuous self-prompt at $2^{\text{nd}}$ and $3^{\text{rd}}$ prompts outperforms the baseline which uses ground-truth (GT) to participate in forming the prompt like the process of the training stage.
However, GT is not available in the real world.
More detailed explanations are presented in Sec.~\ref{sec:Analysis and Discussion}.

\begin{table*}[t]
\vspace{-3mm}
\tablestyle{5.25pt}{1}
\begin{tabular}{l|l|cccccccc|cc}
 \multicolumn{2}{c}{Methods}& \scriptsize GridNet~\cite{grid_dehaze_liu} & \scriptsize PFDN~\cite{pfdn_eccv20_dong}&\scriptsize UHD~\cite{Zheng_uhd_CVPR21}&\scriptsize PSD~\cite{psd_Chen_2021_CVPR}& \scriptsize Uformer~\cite{Wang_2022_CVPR}&\scriptsize Restormer~\cite{Zamir2021Restormer} &\scriptsize D4~\cite{d4_dehze}& \scriptsize Dehamer~\cite{guo2022dehamer} & \scriptsize $\textbf{Ours}_{1}$ & \scriptsize $\textbf{Ours}_{3}$
\vspace{-0.4mm}
\\
 \multicolumn{2}{c}{Venues}&\scriptsize ICCV'19 &\scriptsize ECCV'20 &\scriptsize CVPR'21&\scriptsize CVPR'21 &\scriptsize CVPR'22&\scriptsize CVPR'22 &\scriptsize CVPR'22&\scriptsize CVPR'22& -& -
 \vspace{-0.4mm}
\\
\shline
\scriptsize \multirow{3}{*}{Perception}&NIQE~$\downarrow$ &4.239&4.412&4.743&4.828&4.378&4.321&4.326 &4.529&4.252 & \textcolor{red}{\textbf{4.054}}
\vspace{-0.4mm}
\\
&PI~$\downarrow$ &3.889&4.143&4.962&4.567&3.967&3.936&3.866&  4.035&3.926&\textcolor{red}{\textbf{3.857 }}
\vspace{-0.4mm}
\\
&PIQE~$\downarrow$ &28.924&32.157&39.204&35.174&29.806&29.384& 30.480&32.446&30.596&\textcolor{red}{\textbf{27.927 }}
\vspace{-0.4mm}
\\
\shline
\scriptsize \multirow{2}{*}{\gc{Distortion}}&\gc{PSNR~$\uparrow$} &\gc{32.306}&\gc{33.243}&\gc{16.920}&\gc{13.934}&\gc{33.947} &\gc{36.979}&\gc{19.142} &\gc{36.600}&\gc{35.960} &\gc{34.467}
\vspace{-0.4mm}
\\
&\gc{SSIM~$\uparrow$} &\gc{0.9840}&\gc{0.9827}&\gc{0.7831}&\gc{0.7160}&\gc{0.9846}&\gc{0.9900}&\gc{0.8520}&\gc{0.9865}&\gc{0.9877} &\gc{0.9852}
\end{tabular}
\vspace{-4mm}
\caption{\textbf{Comparisons on the SOTS-indoor dataset.}
Our method achieves better performance in terms of NIQE, PI, and PIQE,
The best results are marked in \textcolor{red}{\textbf{red}}.
$\downarrow$ ($\uparrow$) denotes lower (higher) is better.
$\text{Ours}_{\text{i}}$ means the $i^{\text{th}}$ prompt results.
}
\label{tab:Comparisons with recent SOTAs on the SOTS-indoor dataset.}
\vspace{-2mm}
\end{table*}

\begin{figure*}[!t]
\footnotesize
\centering
\begin{center}
\begin{tabular}{ccccccccc}
\includegraphics[width=0.9999999999999999999\linewidth]{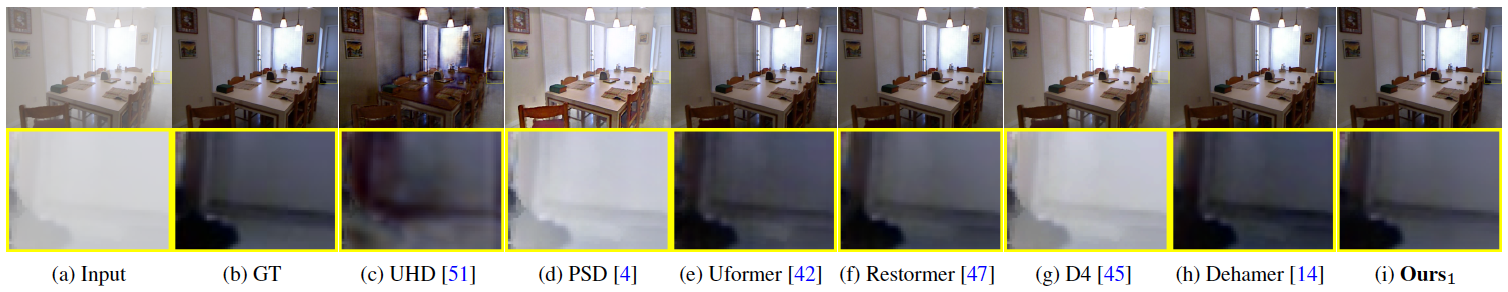}
\end{tabular}
\vspace{-4mm}
\caption{\textbf{Visual comparisons on the SOTS-indoor dataset}.
Our method is able to generate much clearer results, even than the GT image.
}
\label{fig:Visual comparisons on the SOTS-indoor dataset}
\end{center}
\vspace{-6mm}
\end{figure*}
\begin{table*}[!t]
\tablestyle{5.65pt}{1}
\begin{tabular}{l|l|cccccccc|cc}
 \multicolumn{2}{c}{Methods}& \scriptsize GridNet~\cite{grid_dehaze_liu} & \scriptsize PFDN~\cite{pfdn_eccv20_dong}&\scriptsize UHD~\cite{Zheng_uhd_CVPR21}&\scriptsize PSD~\cite{psd_Chen_2021_CVPR}& \scriptsize Uformer~\cite{Wang_2022_CVPR}&\scriptsize Restormer~\cite{Zamir2021Restormer} &\scriptsize D4~\cite{d4_dehze}& \scriptsize Dehamer~\cite{guo2022dehamer} & \scriptsize $\textbf{Ours}_{1}$ & \scriptsize $\textbf{Ours}_{3}$
 \vspace{-0.4mm}
\\
\shline
\scriptsize \multirow{3}{*}{Perception}&NIQE~$\downarrow$ &2.844&2.843&3.756&2.884&2.903&2.956&2.917&3.164 &\textcolor{red}{\textbf{2.646}}&2.685 \\
&PI~$\downarrow$ &2.070&2.326&3.381&2.392&2.241&2.254&2.137&2.251&\textcolor{red}{\textbf{2.003}}&2.027 \\
&PIQE~$\downarrow$ &6.547&6.732&10.891&8.937&6.748&6.904&7.567&6.458&6.577& \textcolor{red}{\textbf{6.151}} \\
\shline
\scriptsize \multirow{2}{*}{\gc{Distortion}}&\gc{PSNR~$\uparrow$} &\gc{16.327}&\gc{16.872}&\gc{11.758}&\gc{15.514}&\gc{19.618}&\gc{18.337} &\gc{26.138}&\gc{ 21.389} &\gc{18.471}&\gc{16.954} \\
&\gc{SSIM~$\uparrow$} &\gc{0.8016}&\gc{0.8532}&\gc{0.6074}&\gc{0.7488}&\gc{0.8798}&\gc{0.8634}&\gc{0.9540}&\gc{0.8926}&\gc{0.8771}&\gc{0.8288} \\
\end{tabular}
\vspace{-3.5mm}
\caption{\textbf{Comparisons on the SOTS-outdoor dataset.}
Our method achieves better perception metrics including NIQE, PI, and PIQE, suggesting that the proposed method has a better generalization ability to unseen images for more natural results generation.
}
\label{tab:Comparisons with recent SOTAs on the SOTS-outdoor dataset.}
\vspace{-3mm}
\end{table*}

\begin{figure*}[!t]
\footnotesize
\centering
\begin{center}
\begin{tabular}{ccccccccc}
\includegraphics[width=0.99999999\linewidth]{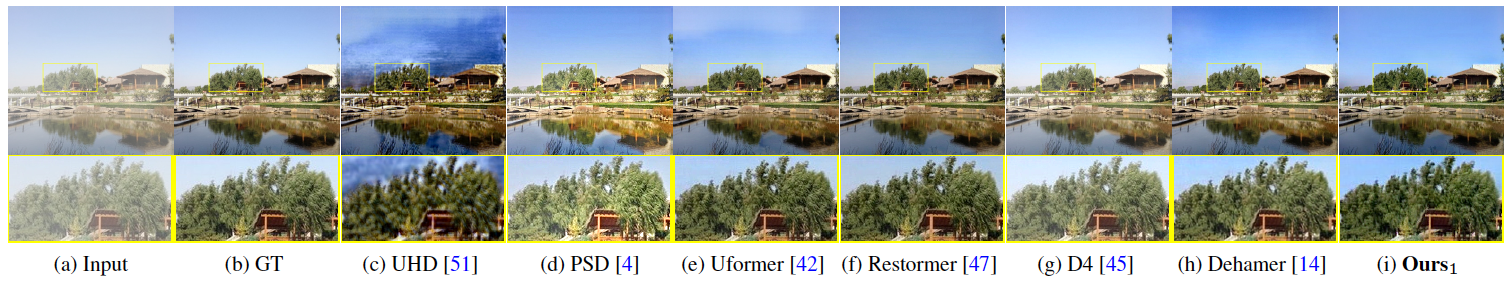}
\end{tabular}
\vspace{-4mm}
\caption{\textbf{Visual comparisons on the SOTS-outdoor dataset}.
Our method is able to generate more natural results.
Note that our method produces more consistent colors in the sky region, while the others generate inconsistent colors and the D4~\cite{d4_dehze} leaves extensive haze.
}
\label{fig:Visual comparisons on the SOTS-outdoor dataset}
\end{center}
\vspace{-10mm}
\end{figure*}
\section{Experimental Results}\label{sec:Experiments}
In this section, we evaluate the effectiveness of our method against state-of-the-art ones (SOTAs) on commonly used benchmarks and illustrate the effectiveness of the key components in the proposed method.
\\
\textbf{Implementation Details.}
We use 10 PTBs, \ie, $l=10$, in our model.
The details about the VQGAN are presented in the supplementary materials.
We crop an image patch of $256\times256$ pixels. The batch size is $10$.
We use ADAM~\cite{adam} with default parameters as the optimizer.
The initial learning rate is 0.0001 and is divided by $2$ at $160$K, $320$K, and $400$K iterations. The model training terminates after $500$K iterations.
The weight parameters $\lambda_{\text{code}}, \lambda_{\text{per}}, \lambda_{\text{adv}}$, and $\lambda_{\text{ssim}}$ are empirically set as 1, 1, 0.1, and 0.5.
Our implementation is based on the PyTorch using one Nvidia 3090 GPU.
\\
\textbf{Synthetic Datasets.}
Following the protocol of~\cite{d4_dehze}, we use the RESIDE ITS~\cite{RESIDE_dehazingbenchmarking_tip2019} as our training dataset and the SOTS-indoor~\cite{RESIDE_dehazingbenchmarking_tip2019} and SOTS-outdoor~\cite{RESIDE_dehazingbenchmarking_tip2019} as the testing datasets.
\\
\textbf{Real-world Datasets.}
In \cite{RESIDE_dehazingbenchmarking_tip2019}, Li~\etal. also collect large-scale real-world hazy images, called UnannotatedHazyImages.
%
We use these images as a real-world hazy dataset.
\\
\textbf{Evaluation Metrics.}
As we mainly aim to recover images with better perception quality, we use widely-used Natural Image Quality Evaluator (\textbf{NIQE})~\cite{niqe}, Perceptual Indexes (\textbf{PI})~\cite{pi}, and Perception-based Image Quality Evaluator (\textbf{PIQE})~\cite{piqe} to measure restoration quality.
Since the distortion metrics Peak-Signal-to-Noise-Ratio (PSNR)~\cite{PSNR_thu} and Structural SIMilarity (SSIM)~\cite{SSIM_wang} cannot model the perception quality well, we use them for  reference only.
Notice that all metrics are re-computed for fairness.
We use the grayscale image to compute the PSNR and SSIM.
We compute NIQE and PI by the provided metrics at https://pypi.org/project/pyiqa/.
The PIQE is computed via https://github.com/buyizhiyou/NRVQA.
%
\begin{table*}[t]
\vspace{-3mm}
\tablestyle{5.85pt}{1}
\begin{tabular}{l|l|cccccccc|cc}
 \multicolumn{2}{c}{Methods}& \scriptsize GridNet~\cite{grid_dehaze_liu} & \scriptsize PFDN~\cite{pfdn_eccv20_dong}&\scriptsize UHD~\cite{Zheng_uhd_CVPR21}&\scriptsize PSD~\cite{psd_Chen_2021_CVPR}& \scriptsize Uformer~\cite{Wang_2022_CVPR}&\scriptsize Restormer~\cite{Zamir2021Restormer} &\scriptsize D4~\cite{d4_dehze}& \scriptsize Dehamer~\cite{guo2022dehamer} & \scriptsize $\textbf{Ours}_{1}$ & \scriptsize $\textbf{Ours}_{3}$
 \vspace{-0.4mm}
\\
\shline
\scriptsize \multirow{3}{*}{Perception}&NIQE~$\downarrow$ &\scriptsize 4.341&\scriptsize4.917&\scriptsize4.515&\scriptsize4.199&\scriptsize4.214 &\scriptsize 4.213&\scriptsize4.257 &\scriptsize4.248&\scriptsize 4.161 &\scriptsize \textcolor{red}{\textbf{4.062}}
\vspace{-0.4mm}
\\
&PI~$\downarrow$ &\scriptsize3.685&\scriptsize3.736&\scriptsize3.858&\scriptsize3.521&\scriptsize3.429&\scriptsize3.436&\scriptsize3.414&\scriptsize3.495&\scriptsize 3.477 &\scriptsize \textcolor{red}{\textbf{3.391}}
\vspace{-0.4mm}
\\
&PIQE~$\downarrow$ &\scriptsize 14.699&\scriptsize17.874&\scriptsize 23.168&\scriptsize 15.851&\scriptsize 16.787&\scriptsize 17.176&\scriptsize  18.678&\scriptsize 15.909&\scriptsize 16.252&\scriptsize \textcolor{red}{\textbf{14.026}}
\end{tabular}
\vspace{-3mm}
\caption{\textbf{Comparisons on the real-world dataset.}
Our method achieves better performance, indicating that our method is more robust to real-world scenarios for realistic results generation.
}
\label{tab:Comparisons with recent SOTAs on the real-world dataset.}
\vspace{-3mm}

\end{table*}
\begin{figure*}[!t]
\footnotesize
\centering
\begin{center}
\begin{tabular}{ccccccccc}
\includegraphics[width=0.999999\linewidth]{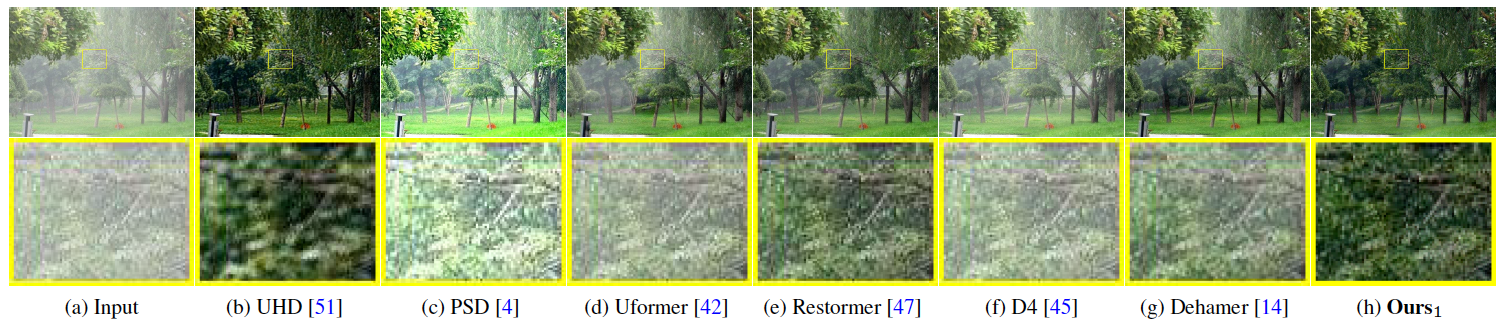}
\end{tabular}
\vspace{-3mm}
\caption{\textbf{Visual comparisons on the real-world dataset}.
Our method generates much clearer results.
Note that existing SOTAs always leave haze residuals, which may be because these methods cannot effectively perceive haze residuals, thus being hard to remove them.
}
\label{fig:Visual comparisons on the real-world dataset}
\end{center}
\vspace{-8mm}
\end{figure*}
\subsection{Results on Synthetic Datasets}
Tab.~\ref{tab:Comparisons with recent SOTAs on the SOTS-indoor dataset.} and Tab.~\ref{tab:Comparisons with recent SOTAs on the SOTS-outdoor dataset.} respectively report the comparison results with SOTAs on the SOTS-indoor and SOTS-outdoor datasets~\cite{RESIDE_dehazingbenchmarking_tip2019}.
Our method achieves better performance in terms of NIQE, PI, and PIQE, indicating the generated results by our method possesses higher perception quality.
Fig.~\ref{fig:Visual comparisons on the SOTS-indoor dataset} and Fig.~\ref{fig:Visual comparisons on the SOTS-outdoor dataset} show that our method restores much clearer images while the evaluated approaches generate the results with haze residual or artifacts.

As we train the network with a one-time feed-forward process, PSNRs and SSIMs are naturally decreased ($\textbf{Ours}_{1}$ \vs $\textbf{Ours}_{3}$ in Tabs.~\ref{tab:Comparisons with recent SOTAs on the SOTS-indoor dataset.} and \ref{tab:Comparisons with recent SOTAs on the SOTS-outdoor dataset.}) when inference is conducted iteratively.
We argue distortion metrics including PSNRs and SSIMs are not good measures for image dehazing as Figs.~\ref{fig:Visual comparisons on the SOTS-indoor dataset} and \ref{fig:Visual comparisons on the SOTS-outdoor dataset} have shown methods with higher PSNR and SSIMs cannot recover perceptual results, \eg, Dehamer~\cite{guo2022dehamer} and D4~\cite{d4_dehze}, while our method with better perception metrics is able to generate more realistic results.
\subsection{Results on Real-World Datasets}
Tab.~\ref{tab:Comparisons with recent SOTAs on the real-world dataset.} summarises the comparison results on the real-world datasets~\cite{RESIDE_dehazingbenchmarking_tip2019}, where our method performs better than the evaluated methods.
Fig.~\ref{fig:Visual comparisons on the real-world dataset} illustrates that our method generates an image with vivid color and finer details.
\subsection{Analysis and Discussion}\label{sec:Analysis and Discussion}
We further analyze the effectiveness of the proposed method and understand how it works on image dehazing.
The results in this section are obtained from the SOTS-indoor dataset if not further mentioned. Our results are from the $1^{\text{st}}$ prompt inference for fair comparisons, \ie, $i=1$ in \eqref{eq: Self-prompt inference} if not further specifically mentioned.
\begin{table}[!t]
\tablestyle{0.25pt}{1.05}
\begin{tabular}{l|ccccccccc}
\scriptsize Experiments &\scriptsize NIQE~$\downarrow$& \scriptsize PI~$\downarrow$ & \scriptsize PIQE~$\downarrow$  & \scriptsize \gc{PSNR~$\uparrow$} &\scriptsize \gc{SSIM~$\uparrow$}
\vspace{-0.4mm}
\\
\shline
\scriptsize \textcolor{blue}{(a)}~Without the prompt &\scriptsize4.258&\scriptsize3.937&\scriptsize31.904&\scriptsize\gc{35.353}&\scriptsize\gc{0.9874}
\vspace{-0.4mm}
\\
\scriptsize \textcolor{blue}{(b)}~Image-level depth difference  &\scriptsize4.901 &\scriptsize 4.343 &\scriptsize 32.141 &\scriptsize \gc{34.573} &\scriptsize \gc{0.9742}
\vspace{-0.4mm}
\\
\scriptsize \textcolor{blue}{(c)}~Concatenation of image and depth features  &\scriptsize 4.362 &\scriptsize 4.077 &\scriptsize 34.107 &\scriptsize \gc{24.378} &\scriptsize \gc{0.9394}
\vspace{-0.4mm}
\\
\shline
\scriptsize  \textcolor{blue}{(d)}~Feature-level depth difference (\textbf{Ours}) &\scriptsize 4.252  &\scriptsize 3.926 &\scriptsize 30.596 &\scriptsize \gc{35.960} &\scriptsize \gc{0.9877}
\end{tabular}
\vspace{-3mm}
\caption{\textbf{Effect of the proposed prompt on image dehazing.}
Feature-level depth difference is a better prompt formalization.
}
\label{tab:Formalization of the prompt.}
\vspace{-2mm}
\end{table}
\begin{figure}[!t]
\footnotesize
\centering
\begin{center}
\begin{tabular}{ccc}
\includegraphics[width=0.9999\linewidth]{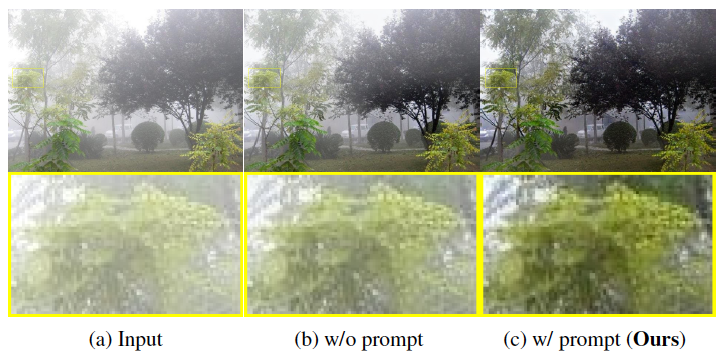}
\end{tabular}
\vspace{-3mm}
\caption{\textbf{Visual comparisons of the model without prompt (b) and with prompt (c) on real-world scenarios}.
Using the prompt helps generate much clearer results.
}
\label{fig: Visual comparisons of the model with w/o prompt and w/ prompt on one real-world hazy image}
\end{center}
\vspace{-8mm}
\end{figure}
\\
{\bf Effectiveness of prompt.}
%
Initially, we assess the effect of the prompt on image dehazing.
Notably, various prospective prompt candidates exist, such as image-level depth difference as the input of the VQGAN encoder or concatenation between deep features extracted from the input and depth features as the input of the Transformers.
Our proposed prompt is compared with these candidates, as illustrated in Tab.~\ref{tab:Formalization of the prompt.}(b) and \ref{tab:Formalization of the prompt.}(c), demonstrating that none of these candidates outperforms our proposed prompt.

\begin{figure}[!t]
\vspace{-2mm}
\footnotesize
\centering
\begin{center}
\begin{tabular}{c}
\hspace{-2mm}\includegraphics[width=0.999\linewidth]{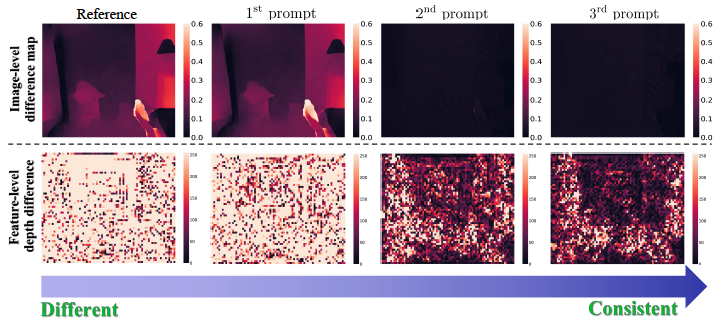}
\end{tabular}
\vspace{-4mm}
\caption{\textbf{Illustration of continuous depth-consistency}.
With the steps of the number of prompts increasing, the depth difference in both image-level and feature-level becomes more consistent.
Reference means the depth difference between the input hazy image and GT.
The input haze image is Fig.~\ref{fig:Visualization on feature-level depth difference}(a).
}
\label{fig: Illustration of depth-consistency}
\end{center}
\vspace{-5mm}
\end{figure}

Note our method without prompt leads to a similar model with CodeFormer~\cite{codeformer} which directly inserts regular Transformers into VQGAN.
Tab.~\ref{tab:Formalization of the prompt.} shows prompt help yield superior perception quality than the model without prompt (Tab.~\ref{tab:Formalization of the prompt.}(a)).
The efficacy of our model with the prompt is further affirmed by Fig.~\ref{fig: Visual comparisons of the model with w/o prompt and w/ prompt on one real-world hazy image}, indicating that the model with the prompt generates better results, while the model without prompt fails to remove haze effectively.
\\
{\bf Effectiveness of prompt embedding.}
One might ponder the relative efficacy of our prompt embedding in contrast to the prevalent technique of position embedding (Fig.~\ref{fig:Existing attention Prompt-based attention}(a)).
In this regard, we assess the effect of these embedding approaches in Tab.~\ref{tab:Effectiveness of prompt embedding.}. The table reveals that our prompt embedding proves more advantageous over the position embedding, since the former is associated with haze residual information.
\begin{table}[!t]
\tablestyle{3.5pt}{1.05}
\begin{tabular}{l|ccccccccc}
\scriptsize Experiments &\scriptsize NIQE~$\downarrow$& \scriptsize PI~$\downarrow$ & \scriptsize PIQE~$\downarrow$  & \scriptsize \gc{PSNR~$\uparrow$} &\scriptsize \gc{SSIM~$\uparrow$}
\vspace{-0.4mm}
\\
\shline
\scriptsize\textcolor{blue}{(a)}~Without embedding&\scriptsize4.410&\scriptsize4.113&\scriptsize32.193 &\scriptsize\gc{34.997}&\scriptsize\gc{0.9704}
\vspace{-0.4mm}
\\
\scriptsize \textcolor{blue}{(b)}~Position embedding&\scriptsize4.267&\scriptsize3.992&\scriptsize31.877&\scriptsize\gc{35.331}&\scriptsize \gc{0.9782}
\vspace{-0.4mm}
\\
\shline
\scriptsize \textcolor{blue}{(c)}~Prompt embedding (\textbf{Ours})&\scriptsize 4.252  &\scriptsize 3.926 &\scriptsize 30.596 &\scriptsize \gc{35.960} &\scriptsize \gc{0.9877}
\end{tabular}
\vspace{-3mm}
\caption{\textbf{Effectiveness of prompt embedding.}
The proposed prompt embedding is better than regular position embedding.
}
\label{tab:Effectiveness of prompt embedding.}
\vspace{-3.5mm}
\end{table}
\begin{table}[!t]
\tablestyle{4.25pt}{1.05}
\begin{tabular}{l|ccccccccc}
\scriptsize Experiments &\scriptsize NIQE~$\downarrow$& \scriptsize PI~$\downarrow$ & \scriptsize PIQE~$\downarrow$  & \scriptsize \gc{PSNR~$\uparrow$} &\scriptsize \gc{SSIM~$\uparrow$}
\vspace{-0.4mm}
\\
\shline
\scriptsize \textcolor{blue}{(a)}~Regular attention&\scriptsize4.300&\scriptsize4.102&\scriptsize31.486 &\scriptsize\gc{35.874}&\scriptsize\gc{0.9811}
\vspace{-0.4mm}
\\
\shline
\scriptsize \textcolor{blue}{(b)}~Prompt attention (\textbf{Ours}) \scriptsize&\scriptsize 4.252  &\scriptsize 3.926 &\scriptsize 30.596 &\scriptsize \gc{35.960} &\scriptsize \gc{0.9877}
\end{tabular}
\vspace{-3.5mm}
\caption{\textbf{Effectiveness of prompt attention.}
Our prompt attention is more effective than regular attention for haze removal.
}
\label{tab:Effectiveness of prompt attention.}
\vspace{-6mm}
\end{table}
\\
{\bf Effectiveness of prompt attention.}
Analyzing the efficacy of prompt attention proves intriguing.
Tab.~\ref{tab:Effectiveness of prompt attention.} indicates that our prompt attention yields better results as compared to commonly used attention methods (Fig.~\ref{fig:Existing attention Prompt-based attention}(c)).
These findings signify that incorporating prompts in enhancing Query estimation accounts for the haze information, thereby culminating in more effective image dehazing results.
\begin{figure}[!t]
\vspace{-2mm}
\footnotesize
\centering
\begin{center}
\begin{tabular}{c}
\hspace{-2mm}\includegraphics[width=0.9\linewidth]{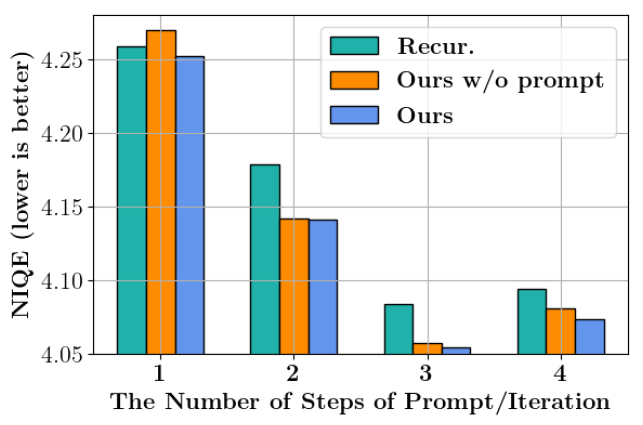}
\end{tabular}
\vspace{-5mm}
\caption{\textbf{Effectiveness of continuous self-prompt (Ours) \vs. recurrent dehazing (Recur.)}.
Our continuous self-prompt inference can further improve results towards better naturalness, and it always produces better results than recurrent dehazing.
'Ours w/o prompt' means the results of \eqref{eq: Step 1}.
}
\label{fig: Effectiveness of continuous self-prompt recurrent dehazing}
\end{center}
\vspace{-4mm}
\end{figure}
\begin{figure}[!t]
\footnotesize
\centering
\begin{center}
\begin{tabular}{cccc}
\includegraphics[width=0.999999\linewidth]{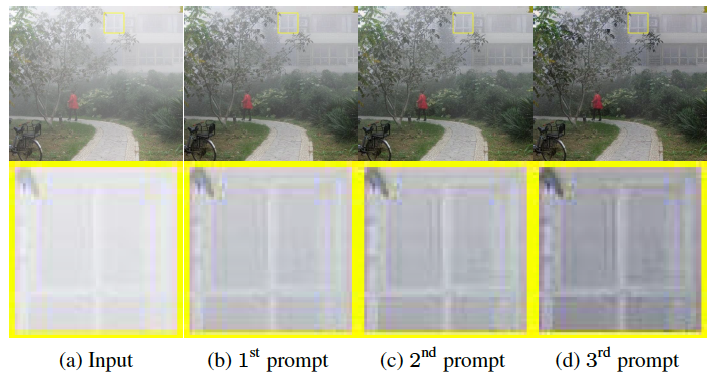}
\end{tabular}
\vspace{-3mm}
\caption{\textbf{Visual improvement of continuous self-prompt inference on a real-world example}.
Our continuous self-prompt inference can progressively improve the dehazing performance.}
\label{fig: Visual improvement of continuous self-prompt inference on real-world scenarios}
\end{center}
\vspace{-6mm}
\end{figure}
\\
{\bf Effect of the number of steps in continuous self-prompt.}
The inference stage involves several steps to generate the prompt for better image dehazing.
We thus examine the effect of the number of steps in the continuous self-prompt.
Fig.~\ref{fig: Effectiveness of continuous self-prompt recurrent dehazing} reveals that the optimal performance is achieved with a number of steps equal to $3$ in the continuous self-prompts (\ie, $i=3$ in \eqref{eq: Self-prompt inference}), in terms of NIQE.
Notably, additional prompts do not improve the dehazing performance any further.
One real-world example in Fig.~\ref{fig: Visual improvement of continuous self-prompt inference on real-world scenarios} demonstrates that our continuous self-prompt method can gradually enhance dehazing quality.
\\
{\bf Continuous self-prompt \vs recurrent dehazing.}
We use the continuous self-prompt approach to restore clear images progressively at inference.
To determine whether a recurrent method that is training our model without prompt achieves similar or better results, we compare our proposed method with it in Fig.~\ref{fig: Effectiveness of continuous self-prompt recurrent dehazing}, demonstrating that the recurrent method is not as good as our continuous self-prompt.
\\
{\bf Continuous self-prompt \vs GT guidance.}
Fig.~\ref{fig: Continuous self-prompt vs GT guidance.} compares the NIQE performance of ground truth (GT) guidance with that of the continuous self-prompt algorithm.
Results show that while GT guidance performs better than the $1^{\text{st}}$ prompt, it falls short of the effectiveness of the $2^{\text{nd}}$ and $3^{\text{rd}}$ prompts.
This is likely due to GT guidance's limited ability to handle haze residuals which may still exist in the dehazed images, which are addressed by the self-prompt's ability to exploit residual haze information to progressively improve dehazing quality over time.
Moreover, as GT is not available in the real world, these findings may further support the use of self-prompt as a more practical alternative.
\\
{\bf Depth-consistency.}
Fig.~\ref{fig: Illustration of depth-consistency} shows heat maps of depth differences obtained by the continuous self-prompt inference with different prompt steps.
The results demonstrate both image-level and feature-level depth differences decrease as the number of prompt steps increases, indicating the depths obtained with the prompt, \ie, \eqref{eq: Step 3}, become increasingly consistent with those obtained without it, \ie, \eqref{eq: Step 1}.
\\
{\bf Model size and running time.}
Tab.~\ref{tab:Model sizes and running speed} compares our model sizes and running time against recent Transformer-based SOTAs: Uformer~\cite{Wang_2022_CVPR} and Dehamer~\cite{guo2022dehamer}.
Our model is comparable with these leading methods on model sizes.
While the single-iteration time speed of our method remains comparable to these two feed-forward models~\cite{Wang_2022_CVPR,guo2022dehamer}, it requires slightly more time for multiple iterations since our method involves estimating depths.


\begin{table}[!t]
\tablestyle{3.25pt}{1}
\begin{tabular}{l|cc|ccc}
\multirow{2}{*}{Methods}&Uformer& Dehamer&\multicolumn{3}{c}{ \scriptsize Ours}
\vspace{-0.4mm}
\\
& \cite{Wang_2022_CVPR}& \cite{guo2022dehamer}& \scriptsize $1^{\text{st}}$ prompt & \scriptsize $2^{\text{nd}}$ prompt & \scriptsize $3^{\text{rd}}$ prompt
\vspace{-0.4mm}
\\
\shline
\scriptsize Parameters&\scriptsize 21M&\scriptsize 132M&\scriptsize 34M&\scriptsize 34M&\scriptsize 34M
\vspace{-0.4mm} 
\\
\scriptsize Running time&\scriptsize 0.15s&\scriptsize 0.16s&\scriptsize 0.19s&\scriptsize 1.32s&\scriptsize 2.41s
\end{tabular}
\vspace{-3mm}
\caption{\textbf{Model sizes and running time.}
The running time is reported on an image with $460\times620$ pixels on one 3090 GPU.
}
\label{tab:Model sizes and running speed}
\vspace{-5mm}
\end{table}
%
\section{Conclusion}\label{sec:Conclusion}
We have proposed a simple yet effective self-prompt Transformer for image dehazing by exploring the prompt built on the estimated depth difference between the image with haze residuals and its clear counterpart.
We have shown that the proposed prompt can guide the deep model for better image dehazing.
To generate better dehazing images at the inference stage, we have proposed continuous self-prompt inference, where the proposed prompt strategy can remove haze progressively.
We have shown that our method generates results with better perception quality in terms of NIQE, PI, and PIQE.
\\
\textbf{Limitations.}
Our model is influenced by the estimated depth of $\Bar{\text{J}}_{i-1}^{\text{w/o~prompt}}$ and $\Bar{\text{J}}_{i}^{\text{w/o~prompt}}$ in \eqref{eq: Step 1}.
Our continuous self-prompt approach may not work well if the depth difference is not significant enough.
Slight performance degradation occurs when multiple prompts in the SOTS-outdoor in terms of NIQE and PI (Tab.~\ref{tab:Comparisons with recent SOTAs on the SOTS-outdoor dataset.}).
We argue this may be because the depth generated from $\Bar{\text{J}}_{i-1}^{\text{w/o~prompt}}$ and $\Bar{\text{J}}_{i}^{\text{w/o~prompt}}$ in \eqref{eq: Step 1} in SOTS-outdoor is not significant enough.
Yet, the $1^{\text{st}}$ prompt still outperforms existing methods.
{\small
\bibliographystyle{ieee_fullname}
\bibliography{egbib}
}

\end{document}